\documentclass[conference]{IEEEtran}
\IEEEoverridecommandlockouts

\RequirePackage[absolute]{textpos}
\DeclareRobustCommand*{\copyrightnote}{%
  \begin{textblock}{85}(17.5,259.5)
      \scriptsize{\noindent \copyright 2021 IEEE. Personal use of this material is permitted. Permission from IEEE must be obtained for all other uses, in any current or future media, including reprinting/republishing this material for advertising or promotional purposes, creating new collective works, for resale or redistribution to servers or lists, or reuse of any copyrighted
component of this work in other works.}%
  \end{textblock}%
    }

\usepackage[
    colorlinks=true,
    linkcolor={blue!65!black},
    citecolor={blue!65!black},
    filecolor=black,
    urlcolor={blue!65!black},
    plainpages=false,
    pdfpagelabels,
    breaklinks=true
    ]{hyperref}

\usepackage{amsmath,amssymb,amsfonts}
\usepackage{algorithmic}
\usepackage{graphicx}
\usepackage{textcomp}
\usepackage{xcolor}
\def\BibTeX{{\rm B\kern-.05em{\sc i\kern-.025em b}\kern-.08em
    T\kern-.1667em\lower.7ex\hbox{E}\kern-.125emX}}
    
\usepackage{placeins}

\usepackage{changes}
\usepackage[nameinlink, capitalize]{cleveref}

\usepackage[giveninits=true,sorting=none,style=ieee]{biblatex}
\AtEveryBibitem{\clearfield{pages}} 

\newcommand{\pderivs}[2]{\frac{\partial #1}{\partial #2}}

\newcommand{\stime}[1]{\pmb{t}_{#1}}

\newcommand{\vrel}[1]{\pmb{r}_{#1}}

\newcommand{\sscore}[3]{\vartheta_{#1,#2,#3}}

\newcommand{\dist}[3]{d_{#1}\left(\stime{#2}, \stime{#3} \right)}
\newcommand{\lone}[1]{\| #1 \|}
\newcommand{\spikediff}[2]{\stime{#1} - \stime{#2}}
\newcommand{\reldiff}[2]{#1 - \vrel{#2}}
\newcommand{\A}{\mathrm{A}}
\renewcommand{\S}{\mathrm{S}}
\newcommand{\sign}[1]{\mathrm{sign}\left(#1 \right)}
\newcommand{\heaviside}[1]{\theta\left( #1 \right)}
\newcommand{\taus}{\tau_\mathrm{s}}

\newcommand{\thresh}{u_\mathrm{th}}
\newcommand{\nlifsum}[1]{\sum_{t_{#1} \leq t^*} W_{s,i#1}}
\newcommand{\nlifsumi}[1]{\sum_{t_{#1} \leq t_{s,i}} W_{s,i#1}}
\newcommand{\nlifsumth}[1]{\nlifsum{#1} - \thresh}
\newcommand{\nlifsumith}[1]{\nlifsumi{#1} - \thresh}
\newcommand{\vwerr}[3]{$#1^{+#2}_{-#3}$}
\newcommand{\triple}[3]{\{$#1$,\ \nolinebreak $#2$,\ \nolinebreak $#3$\}}
\newcommand{\citep}[1]{\cite{#1}}
\newcommand{\citet}[1]{\cite{#1}}

\usepackage{nameref}

\begin{document}

\title{SpikE: spike-based embeddings for\\multi-relational graph data}

\author{\IEEEauthorblockN{Dominik Dold,\ Josep Soler Garrido}
\IEEEauthorblockA{\textit{Siemens AI Lab} \\
\textit{Siemens AG Technology}\\
80331 Munich, Germany \\
\{dominik.dold, josep.soler\_garrido\}@siemens.com}
}

\maketitle
\copyrightnote
\thispagestyle{plain}
\pagestyle{plain}
\begin{abstract}
Despite the recent success of reconciling spike-based coding with the error backpropagation algorithm, spiking neural networks are still mostly applied to tasks stemming from sensory processing, operating on traditional data structures like visual or auditory data.
A rich data representation that finds wide application in industry and research is the so-called knowledge graph -- a graph-based structure where entities are depicted as nodes and relations between them as edges.
Complex systems like molecules, social networks and industrial factory systems can be described using the common language of knowledge graphs, allowing the usage of graph embedding algorithms to make context-aware predictions in these information-packed environments. 
We propose a spike-based algorithm where nodes in a graph are represented by single spike times of neuron populations and relations as spike time differences between populations.
Learning such spike-based embeddings only requires knowledge about spike times and spike time differences, compatible with recently proposed frameworks for training spiking neural networks.
The presented model is easily mapped to current neuromorphic hardware systems and thereby moves inference on knowledge graphs into a domain where these architectures thrive, unlocking a promising industrial application area for this technology. 

\end{abstract}

\section{Introduction}\label{sec:introduction}
Recently, spiking neural networks (SNNs) have started to bridge the gap to their widely used cousins, artificial
neural networks, and achieved competitive performances on benchmark tasks like pattern recognition \citep{mostafa2017supervised,zenke2018superspike,neftci2019surrogate,comsa2019temporal,kheradpisheh2019s4nn,göltz2020fast}, probabilistic inference \citep{neftci2014event,petrovici2016stochastic,dold2019noise}
and sequence prediction tasks \citep{bellec2018long,bellec2019biologically}.
One crucial ingredient for this success was the consolidation of the error backpropagation algorithm with SNNs, which had remained an unsolved problem for a long time due to the discontinuous nature of spike generation \citep{zenke2018superspike,bohte2000spikeprop,huh2018gradient,wunderlich2020eventprop}.
However, so far SNNs have mostly been applied to tasks akin to sensory processing like image or audio recognition \citep{cramer2019heidelberg}.
Such input data is inherently well-structured, e.g., the pixels in an image have fixed positions, and applicability is often
limited to a narrow set of tasks that utilize this structure and do not scale well beyond the initial data domain.

\begin{figure}[t]
    \centering
    \includegraphics[width=\columnwidth]{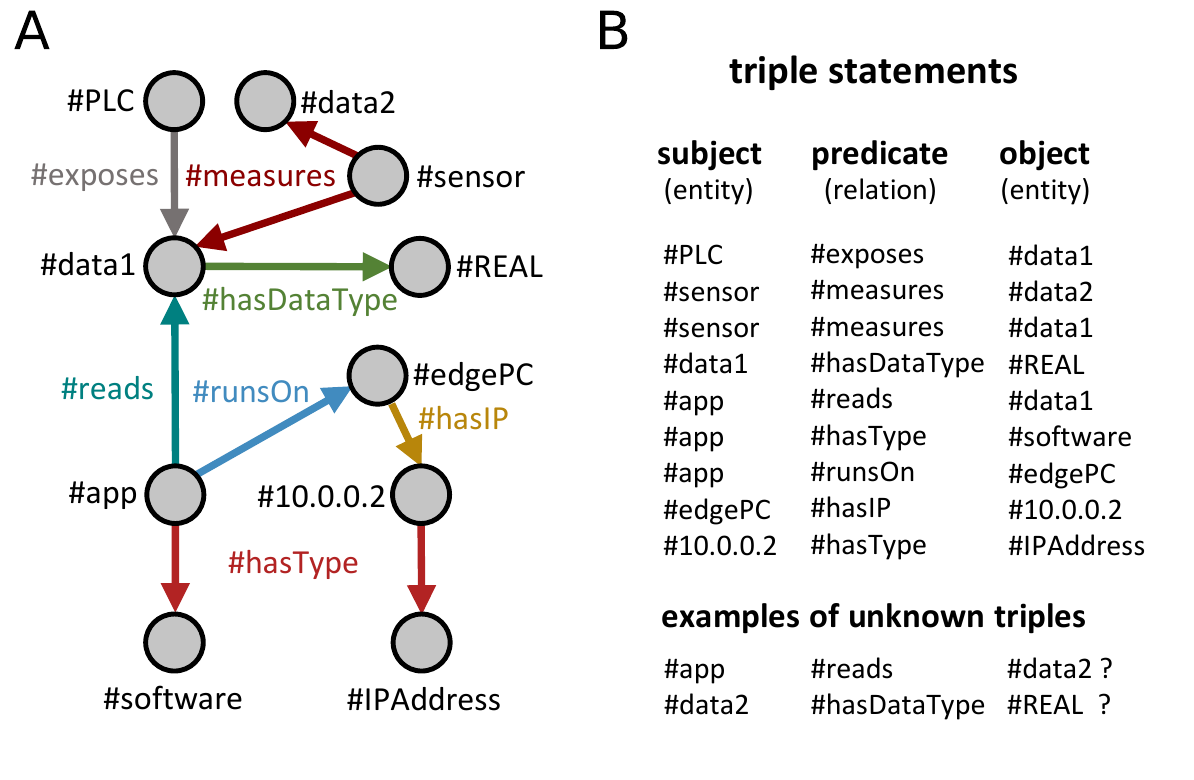}\vspace{-4mm}
	\caption{\textbf{(A)} Simplified example of a graph describing parts of an industrial automation system, with nodes representing entities and edges relations. Relation types are shown in different colors. 
	\textbf{(B)} The graph can be summarized as a set of known triple statements. Inference on graph data is concerned with evaluating whether unknown statements are plausible given the structure of the observed graph.\vspace{-4mm}
	}	
	\label{fig:data}
\end{figure}

A data structure that allows reasoning over abstract concepts and seamless integration of data from different domains are knowledge graphs (KGs) \citep{auer2007dbpedia,bollacker2008freebase,singhal2012introducing}. 
KGs are a widely used, rich data structure that enables a symbolic description of abstract concepts and how they relate to each other.
In general, a KG consists of nodes representing entities and edges representing relations between these entities.
For instance, in an industrial automation system, the nodes could represent physical objects like a sensor or a programmable logic controller (PLC), but also more abstract entities like an IP address, data types or an application running on the industrial system (\cref{fig:data}A).
How these entities relate to each other is modeled with edges of different types between nodes.
This way, the graph can be summarized using semantically meaningful statements, so-called triples, that take the simple and human-readable form \triple{\text{subject}}{\text{predicate}}{\text{object}} \citep{brickley1999resource}, or in graph format, \triple{\text{node}}{\text{typed edge}}{\text{node}} (\cref{fig:data}B).

Although multi-relational graphs are highly expressive, their symbolic nature prevents the direct usage of classical statistical methods for further processing and evaluation. 
Lately, graph embedding algorithms have been introduced to solve this problem by mapping nodes and edges to a vector space while conserving certain graph properties \citep{nickel2015review,hamilton2017representation,ruffinelli2019you}.
For example, one might want to conserve a node's proximity, such that connected nodes or nodes with vastly overlapping neighborhoods are mapped to vectors that are close to each other.
These vector representations can then be used in traditional machine learning approaches to make predictions about unseen statements, realizing abstract reasoning over a set of subjects, predicates and objects.

Spike-based versions of classical graph algorithms like finding shortest paths, minimum spanning trees and maximum flows have already been proposed in recent work \citep{hamilton2017community,hamilton2018towards,hamilton2018neural,schuman2019shortest,ali2019spiking,hamilton2019spike,kay2020neuromorphic}.
We extend this work to graph embeddings for multi-relational graphs, where instead of working directly with the graph structure, it is encoded in the temporal domain of spikes: entities and relations are represented as spikes of neuron populations and spike time differences between populations, respectively.
Through this mapping from graph to spike-based coding, SNNs can be trained on graph data to evaluate novel triples not seen during training, i.e., perform inference on the semantic space spanned by the training graph.
For our studies, we use non-leaky integrate-and-fire neurons (nLIF), allowing us to calculate spike times and spike time gradients analytically \citep{mostafa2017supervised} while guaranteeing compatibility with current neuromorphic hardware architectures \citep{thakur2018large,roy2019towards} that often realize some variant of the LIF neuron model.
The presented results are especially interesting for the applicability of neuromorphic hardware in industrial use-cases \citep{davies2019benchmarks}, where graph embedding algorithms find many applications, e.g., in form of recommendation systems \citep{hildebrandt2018configuration}, digital twins \citep{ringsquandl2017event}, semantic feature selectors \citep{ringsquandl2015semantic} or anomaly detectors \citep{soler2021graph}.

In the following, we first explain our spike-based graph embedding model (SpikE), derive the required learning rule and evaluate the learned embeddings on a realistic industrial benchmark data set.

\section{Spike-based graph embeddings}
\subsection{From graphs to spikes} Our model takes inspiration from TransE \citep{bordes2013translating}, a shallow graph embedding algorithm where nodes are represented as vectors and relations as vector translations.
In principle, we found that these vector representations can be mapped to spike times and translations into spike time differences, offering a natural transition from the graph domain to SNNs.

\begin{figure}[t]
    \centering
    \includegraphics[width=0.9\columnwidth]{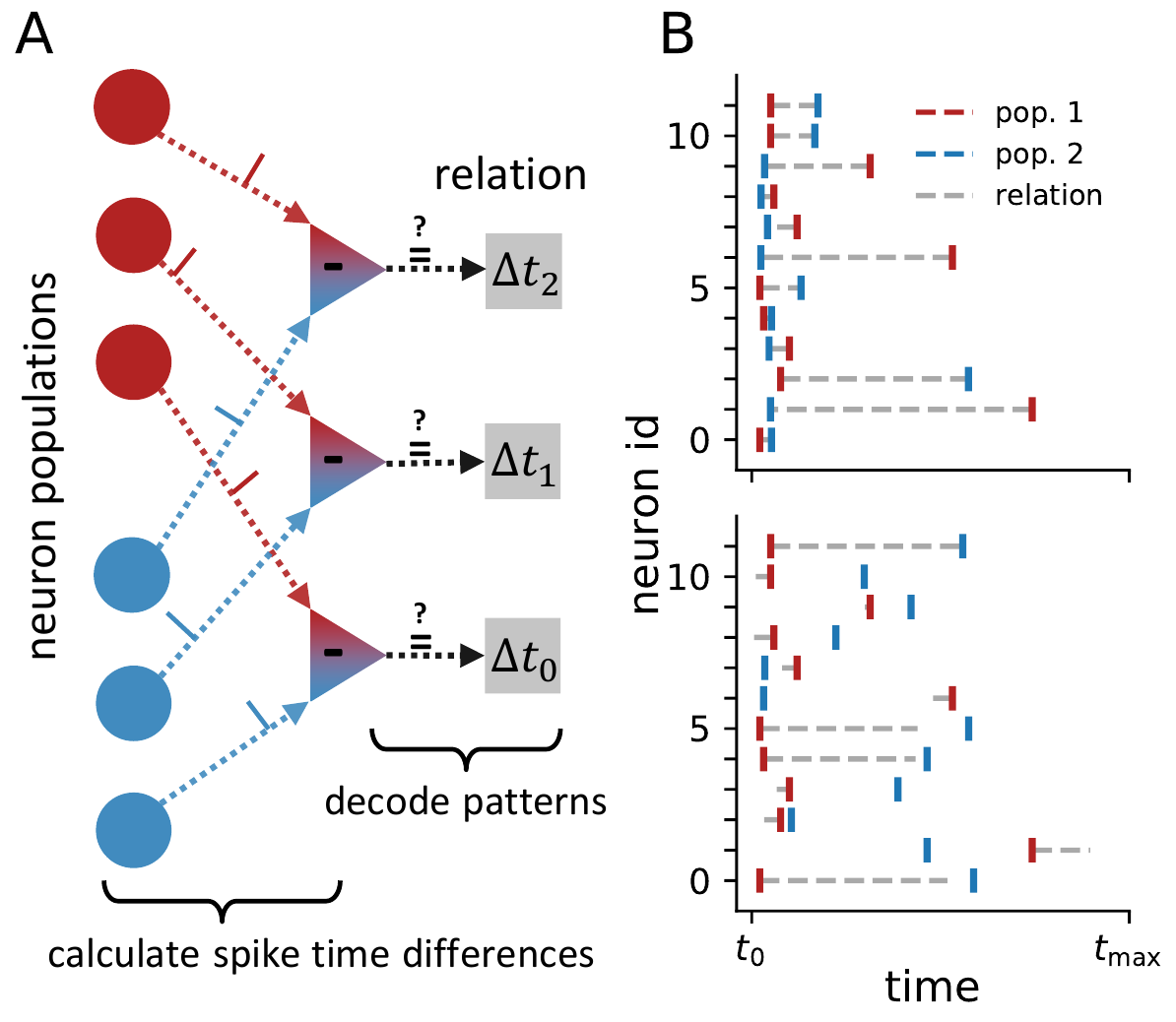}
	\caption{\textbf{(A)} Spike-based coding scheme to embed graphs into SNNs.
	Nodes are represented by neuron populations (red, blue), where the embedding is given by the individual spike time of each neuron.
	By comparing spike time differences between populations (triangles), one can evaluate whether certain relations (gray boxes) are valid between the two entities encoded by the populations.
	\textbf{(B)} Example of spike patterns and spike time differences for a plausible triple (top) and an implausible one (bottom), i.e., where the pattern does not match the relation, pop. 1 $-$ pop. 2 $\neq$ relation. In both cases, we used the same subject (red), but different relations and objects (gray and blue).\vspace{-4mm}
	}	
	\label{fig:coding}
\end{figure}

We propose that the embedding of a node $s$ is given by single spike times of a neuron population of size $N$, $\stime{s} \in [t_0,t_\mathrm{max}]^N \in \mathbb{R}^N$ (\cref{fig:coding}A).
That is, every neuron of the population emits exactly one spike during the time interval $[t_0, t_\mathrm{max}]$, and the resulting spike pattern represents the embedding of an entity in the KG.
Relations are encoded by a $N$-dimensional vector of spike time differences $\vrel{p} \in \mathbb{R}^N$.
To decode whether two populations $s$ and $o$ encode entities that are connected by relation $p$, we evaluate the spike time differences of both populations element-wise, $\stime{s}-\stime{o}$, and compare it to the entries of the relation vector $\vrel{p}$ (\cref{fig:coding}A).
Depending on how far these diverge from each other, the statement \triple{s}{p}{o} is either deemed plausible or implausible.
For instance, in \cref{fig:coding}B (top), the spike pattern of the two populations encoding subject and object entity are consistent with the representation of the relation, i.e., $\stime{s} - \stime{o} \approx \vrel{p}$, and hence the triple \triple{s}{p}{o} is deemed plausible.
In \cref{fig:coding}B (bottom), we choose a triple \triple{s}{q}{b} that is assessed as implausible by our model, since the measured spike time differences do not match those required for relation $q$, i.e., $\stime{s} - \stime{b} \neq \vrel{q}$ (although they might match other relations).

\begin{figure*}[t]
    \centering
    \includegraphics[width=0.9\textwidth]{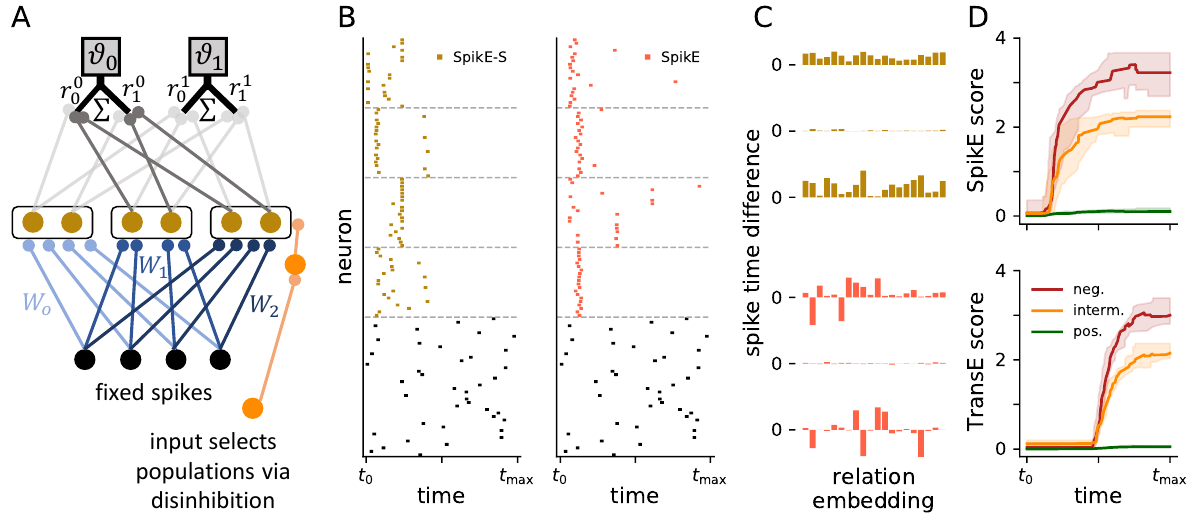}
	\caption{\textbf{(A)} Illustration of the proposed network architecture.
	Fixed spikes (black) and plastic weights (blue) encode the spike times of the embedding populations (gold), which statically project to dendritic compartments of output neurons (gray)\protect\footnotemark.
	To score triples, the adequate populations are activated using, e.g., a disinhibition mechanism (orange)\protect\footnotemark.
	\textbf{(B)} Fixed stimulus spikes (black) and examples of learned spike time embeddings for SpikE-S (gold) and SpikE (red).
	\textbf{(C)} Learned relation embeddings in the output neurons. In case of SpikE-S, only positive spike time differences are learned.
	In both cases, complex spike difference patterns are learned to encode relations as well as simpler ones that mostly rely on coincidence detection (middle), i.e., $\vrel{p} \approx 0$.
	\textbf{(D)} Temporal evaluation of triples \triple{s}{p}{o}, for varying degrees of plausibility of the object: (green) seen during training, (orange) not seen during training, but plausible and (red) least plausible (see also \cref{fig:dataevents}A for a similar experiment). Different to TransE that lacks a concept of time, SpikE prefers embeddings where most neurons spike early, allowing faster evaluation of scores. Lines mark the mean score and shaded areas the 15$^\mathrm{th}$ and 85$^\mathrm{th}$ percentile for 10 random seeds.\vspace{-4mm}}
	\label{fig:spikeraster}
\end{figure*}

This coding scheme maps the rich semantic space of graphs into the spike domain, where the spike patterns of two populations encode how the represented entities relate to each other, but not only for one single relation, but the whole set of relations spanning the semantic space.
To achieve this, learned relations encompass a range of patterns from mere coincidence detection to complex spike time patterns.
In fact, coding of relations as spike coincidence detection does naturally occur as a special case in our model when training SNNs on real data, see for example \cref{fig:spikeraster}C.
Such spike embeddings can either be used directly to predict or evaluate novel triples, or as input to other SNNs that can then utilize the semantic structure encoded in the embeddings for subsequent tasks.
\addtocounter{footnote}{-2}

Formally, the ranking of triples can be written as
\begin{equation}
    \sscore{s}{p}{o} = \sum\lone{\reldiff{\dist{}{s}{o}}{p}} \,,
\end{equation}
where $d: \mathbb{R}^N \times \mathbb{R}^N \to \mathbb{R}^N$ is a difference function and the sum is over vector components.
In the remaining document, we call $\sscore{s}{p}{o}$ the score of triple \triple{s}{p}{o}, where plausible triples have a score close to $0$ and implausible ones $\gg 0$.
For SpikE, we define the difference function to be
\begin{equation}
    \dist{\A}{s}{o} = \spikediff{s}{o} \,,
\end{equation}
where both the order and distance of spike times are used to encode relations.
This can be further modified to only incorporate spike time differences,  
\begin{equation}
    \dist{\S}{s}{o} = \lone{\spikediff{s}{o}} \,,
\end{equation}
such that there is no distinction between subject and object populations.
We call this version of the model SpikE-S, which enables a more compact network realization since the same neuron population can be used to represent an entity as subject and object. 
In contrast, SpikE has the benefit of being closer to the original TransE algorithm by utilizing spike time ordering, but might be harder to realize in neuromorphic hardware (see \cref{fig:spikeraster}A for details).
For the sake of completeness, we investigate both versions of the spike-based embedding model.

\subsection{Network implementation}

A suitable neuron model that suffices the requirements of the presented coding scheme, i.e., single-spike coding and being analytically treatable, is the nLIF neuron model. 
For similar reasons, it has recently been used in hierarchical networks utilizing first-spike latency codes \citep{mostafa2017supervised}.
For the neuron populations encoding entities, we use the nLIF model with an exponential synaptic kernel
\begin{equation}\label{eq:dotu}
    \dot{u}_{s,i}(t) = \frac{1}{\taus}\sum_{j} W_{s,ij} \, \heaviside{t - t_j} \exp\left(-\frac{t-t_j}{\taus}\right) \,,
\end{equation}
where $u_{s,i}$ is the membrane potential of the $i$th neuron of population $s$, $\taus$ the synaptic time constant and $\theta(\cdot)$ the Heaviside function.
A spike is emitted when the membrane potential crosses a threshold value $\thresh$.
$W_{s,ij}$ are synaptic weights from a pre-synaptic neuron population, with every neuron $j$ emitting a single spike at fixed time $t_j$ (\cref{fig:spikeraster}A, bottom).
This way, the coding in both stimulus and embedding layers are consistent with each other and the embedding spike times can be adjusted by changing synaptic weights $W_{s,ij}$.
\cref{eq:dotu} can be solved analytically
\begin{equation}\label{eq:u}
    u_{s,i}(t) = \sum_{t_j \leq t} W_{s,ij} \left[ 1 - \exp \left(-\frac{t-t_j}{\taus} \right) \right] \,,
\end{equation}
which is later used to derive a learning rule for the embedding populations.

For relations, we use output neurons with a similar structure as proposed in \citet{dold2021tensor}.
Each output neuron consists of a 'dendritic tree', where branch $k$ evaluates the $k$th component of the spike pattern difference, i.e., $\lone{\reldiff{\dist{}{s}{o}}{p}}_k$, and the tree structure subsequently sums over all contributions, giving $\sscore{s}{p}{o}$ (\cref{fig:spikeraster}A, top).
This way, the components of $\vrel{p}$ become available to all entity populations, despite being locally stored.

Different from ordinary feedforward or recurrent SNNs, the input is not given by a signal that first has to be translated into spike times and is then fed into the first layer (or specific input neurons) of the network.
Instead, inputs to the network are observed triples \triple{s}{p}{o}, i.e., statements that have been observed to be true.
Since all possible entities are represented as neuron populations, the input simply gates which populations become active (\cref{fig:spikeraster}A, orange), resembling a memory recall.
During training, such recalled memories are then updated to better predict observed triples.
Through this memory mechanism, an entity $s$ can learn about global structures in the graph.
For instance, since the representation of a relation $p$ contains information about other entities that co-occur with it in triples, \triple{m}{p}{n}, $s$ can learn about the embeddings of $m$ and $n$ (and vice versa) -- even if $s$ never appears with $n$ and $m$ in triples together.
\addtocounter{footnote}{-1}
\footnotetext{To ease notation, the upper index denotes relation types and the lower index vector components here.}
\addtocounter{footnote}{+1}
\footnotetext{For SpikE, the order in which the spike time differences are calculated is crucial. This can be achieved by either using a more involved gating mechanism (\nameref{si:gating}), or by representing each entity via subject and object populations that are synchronized during training (\nameref{si:synchronize}).}

\subsection{Learning rules}

To learn spike-based embeddings for entities and relations, we use a soft margin loss
\begin{subequations}
\begin{align}
    l_{s,p,o} = &\log \left[ 1 + \exp \left( \sscore{s}{p}{o} \cdot \eta_{s,p,o} \right) \right] \,, \\
    &L(\pmb{\vartheta}, \pmb{\eta}) = \sum_{s,p,o} l_{s,p,o} \label{eq:loss} \,,
\end{align}
\end{subequations}
where $\eta_{s,p,o} \in \{1, -1\}$ is a modulating teaching signal that establishes whether an observed triple \triple{s}{p}{o} is regarded as plausible ($\eta_{s,p,o} = 1$) or implausible ($\eta_{s,p,o} = -1$).
This is required to avoid collapse to zero-embeddings that simply score all possible triples with $0$.
In the graph embedding literature, implausible (negative) examples are generated by corrupting plausible (positive) triples, i.e., given a training triple \triple{s}{p}{o}, either $s$ or $o$ are randomly replaced -- a procedure called 'negative sampling' \citep{ruffinelli2019you,bordes2013translating}.

The learning rules are derived by minimizing the loss via gradient descent.
In addition, as described in \citet{mostafa2017supervised}, we add a regularization term to the weight learning rule that counters silent neurons (\nameref{si:details}).
The gradient for entities can be separated into a loss-dependent error and a neuron-model-specific term
\begin{equation}
    \pderivs{l_{s,p,o}}{W_{s,ik}} = \pderivs{l_{s,p,o}}{t_{s,i}} \pderivs{t_{s,i}}{W_{s,ik}} \,,
\end{equation}
while the gradient for relations only consists of the error $\pderivs{l_{s,p,o}}{\vrel{p}}$.
The error terms are given by (\nameref{si:spike})
\begin{subequations}
\begin{align}
  \pderivs{l_{s,p,o}}{\stime{s}} &= \epsilon_{s,p,o} \cdot \sign{\dist{\A}{s}{o} - \vrel{p}} \,, \\
  \epsilon_{s,p,o} &=  \eta_{s,p,o} \cdot \sigma \left( \sscore{s}{p}{o} \cdot \eta_{s,p,o} \right) \label{eq:error} \,, \\
  \pderivs{l_{s,p,o}}{\stime{o}} &= \pderivs{l_{s,p,o}}{\vrel{p}} = - \pderivs{l_{s,p,o}}{\stime{s}} \,, \end{align}
\end{subequations}
for SpikE and 
\begin{subequations}
\begin{align}
  \pderivs{l_{s,p,o}}{\stime{s}} &= \epsilon_{s,p,o} \cdot \sign{\spikediff{s}{o}} \sign{\dist{\S}{s}{o} - \vrel{p}} \,, \\
  \pderivs{l_{s,p,o}}{\stime{o}} &= - \pderivs{l_{s,p,o}}{\stime{s}} \,, \\
  \pderivs{l_{s,p,o}}{\vrel{p}} &= -\epsilon_{s,p,o} \cdot \sign{\dist{\S}{s}{o} - \vrel{p}} \,,
\end{align}
\end{subequations}
for SpikE-S, where $\sigma(\cdot)$ is the logistic function.

The neuron-specific term can be evaluated using \cref{eq:u}, resulting in (\nameref{si:spike})
\begin{equation}
    \pderivs{t_{s,i}}{W_{s,ik}} = \frac{\taus \heaviside{t_{s,i} - t_k} \left( e^{\left(t_k - t_{s,i}\right) \slash \taus}  - 1 \right)}{\nlifsumith{j}} \,.
\end{equation}
For relations, all quantities in the update rule are accessible in the output neuron.
Apart from an output error, this is also true for the update rules of nLIF spike times.
Specifically, the learning rules only depend on spike times -- or rather spike time differences -- pre-synaptic weights and neuron-specific constants, compatible with recently proposed learning rules for SNNs \citep{mostafa2017supervised,comsa2019temporal,göltz2020fast}.
Alternative losses like the (pairwise) hinge loss \citep{bordes2013translating} can be used as well to derive learning rules.
\section{Experiments}
\subsection{Data}

\begin{figure}[t]
    \centering
    \includegraphics[width=0.9\columnwidth]{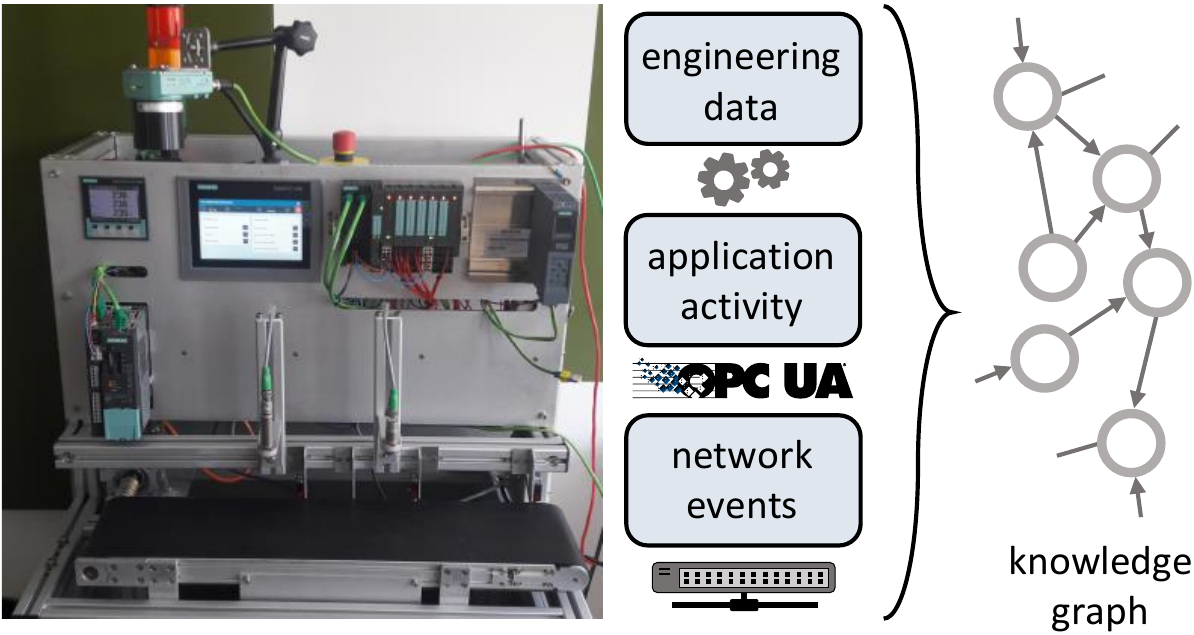}
	\caption{Industrial automation demonstrator (left) used as a data source. Both static engineering data as well as dynamic application activity and network events (middle) are integrated in a KG (right).\vspace{-4mm}}
	\label{fig:demonstrator}
\end{figure}

The proposed model is evaluated on an industrial automation demonstrator (\cref{fig:demonstrator}), designed to capture the complexity of modern industrial automation systems that combine operational technology (OT) components with an information technology (IT) infrastructure.
This convergence of OT and IT technology promises massive improvements in the efficiency and flexibility of industrial manufacturing systems, but comes with challenges like ensuring system integrity and information security \cite{paes2019guide}.
The OT side of our demonstrator has a SIMATIC S7-1500 PLC at its core, connected via an Industrial Ethernet network to multiple subsystems, such as a conveyor belt, industrial cameras or multiple input-output modules connected to sensors. 
The PLC exposes the internal state of some of these subsystems and sensors through a data interface provided by an OPC UA \cite{mahnke2009opc} server. 
This is leveraged on the IT side of the demonstrator, which includes a series of edge computing systems hosting a number of applications, each of which regularly reads or writes data variables on the OPC UA server corresponding to specific parts of the OT system.

The complex set of interactions between the different elements of the demonstrator, e.g., data accesses and network connections, can be naturally represented in a KG and modeled using graph embedding algorithms.
Such a model allows us to evaluate the likelihood of observed system interactions and can be used in anomaly detection tasks to expose unexpected behaviors of the different system components, which may be indicative of a loss of integrity or a cybersecurity incident.

Using such a demonstrator enables us to generate realistic data and benchmarks in a flexible and controllable way \citep{soler2021graph}.
For the following experiments, we use a recording from the demonstrator system with some default network and application activity, resulting in a KG with 3529 nodes, 11 node types, 2 applications, 21 IP addresses, 39 relation types, 360 network events and 472 data access events.
We randomly split the graph with a ratio of 8/2 into mutually exclusive training and test sets, resulting in 12399 training and 2463 test triples.
Details of all experiments can be found in \nameref{si:details}.
For more information on the data generation process, see \citep{soler2021graph}.
\begin{figure*}[ht]
    \centering
    \includegraphics[width=0.9\textwidth]{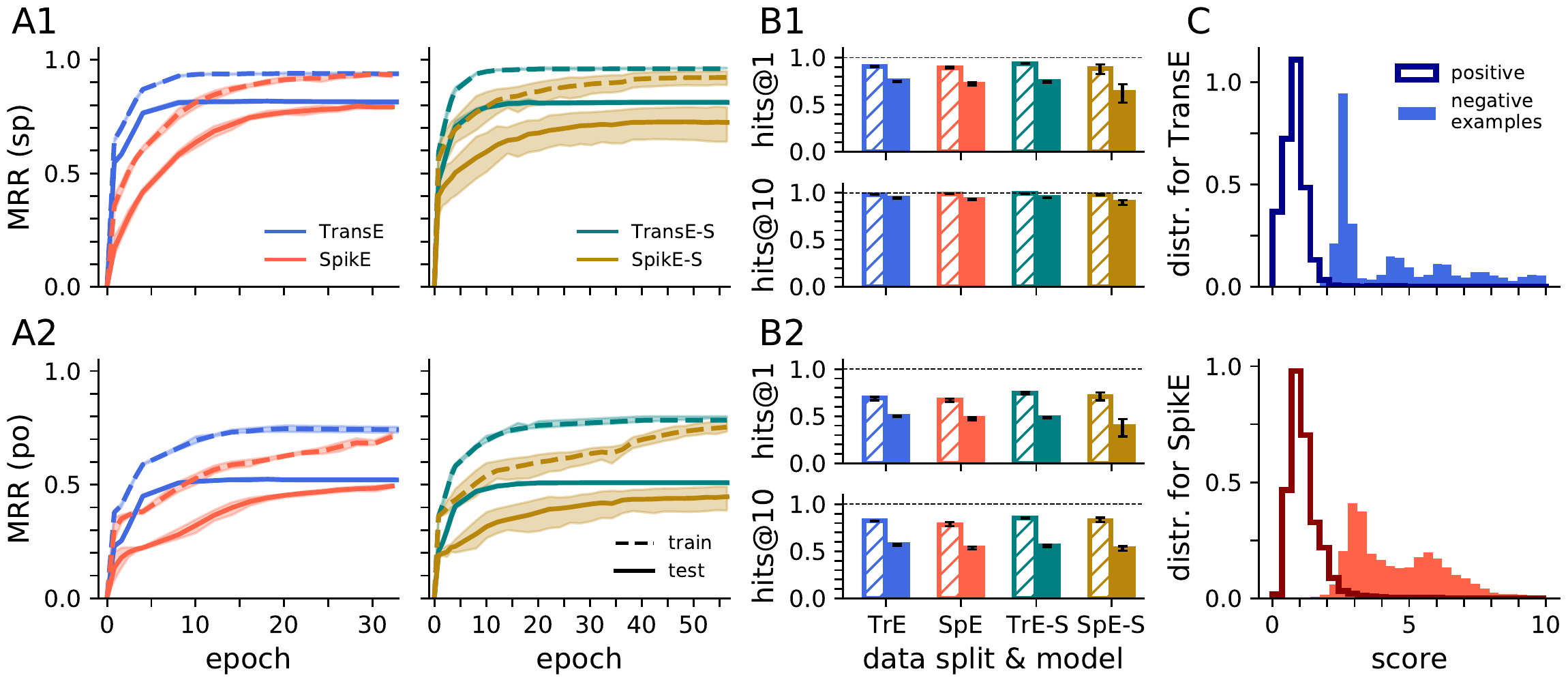}
	\caption{\textbf{(A1)} MRR during training for different models and data splits. The MRR is evaluated for objects, i.e., given a triple \triple{s}{p}{o}, $o$ is ranked against all other possible objects.
	Training was repeated 10 times for different random seeds.
	Shaded areas mark the 15$^\mathrm{th}$ and 85$^\mathrm{th}$ percentile.
	\textbf{(A2)} Same as A1, but for subjects.
	\textbf{(B1)} Mean hits@k scores for objects as in A1.
	Error bars mark the percentiles as in A1 and A2, and train and test data is represented by striped and filled bars, respectively.
	\textbf{(B2)} Same as B1, but for subjects.
	\textbf{(C)} Distribution of scores for positive and negative test triples. As expected, positive examples get scores close to $0$, while negative examples have bad scores, i.e., high mismatch between spike patterns.\vspace{-4mm}}
	\label{fig:MRR}
\end{figure*}
\subsection{Model evaluation}

In \cref{fig:spikeraster}B, C, we showcase some of the learned spike embeddings for entities and relations.
We found that due to the temporal aspect of SNNs -- and the fixed time interval imposed by the stimulus layer -- most embedding neurons spike early.
By evaluating scores over time in an event-based fashion, implausible triples can be identified quickly (\cref{fig:spikeraster}D), which is beneficial when short reaction times are required, e.g., to raise an alarm.
This is not the case for TransE, where embeddings are always symmetric around some baseline offset.

Next, we compare the performance of SpikE and TransE (and their symmetric counterparts SpikE-S and TransE-S) on some traditional graph embedding metrics.
For performance evaluation, we use the mean reciprocal rank (MRR) and hit-based metrics (hits@k) that are often used in the graph embedding literature.
In both cases, a plausible triple \triple{s}{p}{o} is taken and all alternative completions \triple{s}{p}{?} (\cref{fig:MRR}A1, B1) or \triple{?}{p}{o} (\cref{fig:MRR}A2, B2) are scored and sorted.
hits@k measures how frequently the original triple \triple{s}{p}{o} is under the k best scored triples, e.g., hits@1 measures how often it is the top ranked triple compared to all other alternatives.
The MRR averages the reciprocal of the ranks over the presented data set, e.g., if a triple has the second best score (rank 2) it contributes $1/2$ to the MRR.

All models achieve comparable results on these metrics, with a total MRR, i.e., both cases combined, of: \vwerr{0.843}{0.007}{0.007}, \vwerr{0.671}{0.003}{0.003} (TransE); \vwerr{0.872}{0.006}{0.004}, \vwerr{0.661}{0.004}{0.004} (TransE-S); \vwerr{0.824}{0.009}{0.006}, \vwerr{0.645}{0.009}{0.007} (SpikE); \vwerr{0.838}{0.021}{0.020}, \vwerr{0.587}{0.040}{0.050} (SpikE-S), for train and test split (uncertainties are given by the 15$^\mathrm{th}$ and 85$^\mathrm{th}$ percentile).
To ensure that the algorithms are also capable of separating plausible from implausible statements, we show the pure triple scores both 
for the test data and negative examples generated from it (\cref{fig:MRR}C).
\begin{figure}[t]
    \centering
    \includegraphics[width=0.975\columnwidth]{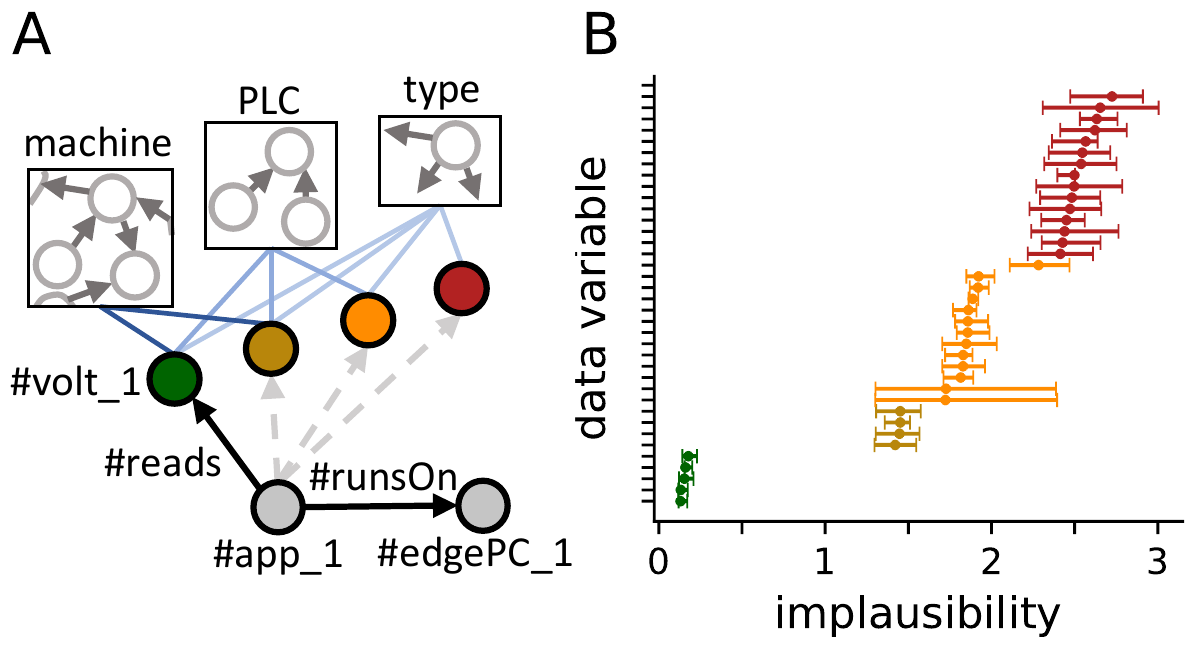}
	\caption{\textbf{(A)} An application reading data from the industrial automation system. 
	There are various ways how data variables accessed during training (green) are related to other data variables in the system.
	For instance, they might (gold) share internal structures documented in the engineering data, (orange) be accessible from the same PLC or (red) only share type-based similarities.
	\textbf{(B)} Accesses to various data variables sorted from top to bottom according to their assigned implausibility (SpikE score). 
	Colors are as in A.
	Bars mark the 15$^\mathrm{th}$ and 85$^\mathrm{th}$ percentile for 10 different random seeds.
	In the background, a second application is active that regularly reads the two data variables with high uncertainty (orange), showing that the embedding of \#app\_1 also learns about the behavior of \#app\_2.\vspace{-4mm}}
	\label{fig:dataevents}
\end{figure}

\subsection{Context-aware decision making}

We further apply SpikE to an anomaly detection task, where an application reads different data variables from the industrial system during training and test time (\cref{fig:dataevents}A).
Data events are sorted according to their SpikE score, with the least plausible data access being on top (\cref{fig:dataevents}B).
As expected, our model utilizes contextual information available through the structure of the KG to rank previously unseen data events, i.e., the less related data variables are to the ones read during training, the higher they end up in the ranking.
For instance, SpikE clearly discerns whether the application accesses variables that are exposed (\cref{fig:dataevents}B, green, gold and orange) or not exposed (red) by the PLC.

\section{Discussion}\label{sec:discussion}
We present a model for spike-based graph embeddings, where nodes and relations of a KG are mapped to spike times and spike time differences in a SNN, respectively.
This allows a natural transition from symbolic elements in a graph to the temporal domain of SNNs, going beyond traditional data formats by enabling the encoding of complex structures into spikes. 
Representations are learned using gradient descent on an output cost function, which yields learning rules that depend on spike times and neuron-specific variables.

In our model, input gates which populations become active and consequently updated by plasticity.
This memory mechanism allows the propagation of knowledge through all neuron populations -- despite the input being isolated triple statements.
After training, the learned embeddings can be used to evaluate or predict arbitrary triples that are covered by the semantic space of the KG.
Moreover, learned spike embeddings can be used as input to other SNNs, providing a native conversion of data into spike-based input -- something that is currently either done artificially by, e.g., mapping pixel values of images to rates or latency codes \citep{pfeiffer2018deep,schuman2019non}, or by restricting oneself to event-based sensors that already provide data in a SNN-compatible format \citep{evanusa2019event,massa2020efficient}.

The nLIF neuron model used in this work is well suited to represent embeddings, but comes with the drawback of a missing leak term, i.e., the neurons are modeled as integrators with infinite memory.
This is critical for neuromorphic implementations, where -- most often -- variations of the nLIF model with leak are realized \citep{thakur2018large}.
Gradient-based optimization of current-based LIF neurons, i.e., nLIF with leak, that extend the results of \citet{mostafa2017supervised} have been demonstrated in
\citet{comsa2019temporal,göltz2020fast}.
Since our model basically exchanges the cost function and network architecture used in \citet{mostafa2017supervised}, these results can be directly applied to our model as well, making it applicable to energy-efficient neuromorphic implementations similar to the ones presented in \citet{göltz2020fast}.
In contrast, output neurons take a simple, but function-specific form that is different from ordinary nLIF neurons.
Although realizable in neuromorphic devices, we believe that alternative forms are possible.
For instance, each output neuron might be represented by a small forward network of spiking neurons \citep{poirazi2003pyramidal}, or relations could be represented by learnable delays \citep{wright2012learning,wang2019delay,zhang2020supervised} or even remain constant (see \nameref{si:additional}).

The presented results bridge the areas of graph analytics and SNNs, opening a new direction in the ongoing endeavor of identifying how SNNs can encode complex information \citep{zenke2021visualizing}.
In future work, this might be extended beyond the single-spike coding scheme, enabling an efficient representation of static graph data by fully utilizing the temporal domain of SNNs, e.g., by encoding nodes as spike trains of individual neurons.
Finally, such models promise exciting and novel industrial applications of event-based neuromorphic devices, e.g., as energy-efficient and flexible processing and learning units for online evaluation of industrial graph data. 
Even though we only hint at the industrial applicability here, more involved benchmarks are currently being investigated in \citep{soler2021graph,dold2021tensor}.

\section*{Acknowledgment}\label{sec:ack}
\addcontentsline{toc}{section}{Acknowledgment}
We thank Marcel Hildebrandt, Serghei Mogoreanu and Martin Ringsquandl for helpful discussions, Johannes Frank for setting up the industrial automation demonstrator as well as Steffen Lamparter, Robert Lohmeyer, Denis Krompaß, Florian Büttner, Mark Buckley, Ulli Waltinger and Benno Blumoser for supporting this work.

\printbibliography
\addcontentsline{toc}{section}{References}

% \vspace{-2.7mm}
\section*{Supplementary Information}
\addtocontents{toc}{\protect\setcounter{tocdepth}{0}}
\subsection[Suppl. A]{Spike-based model}\label{si:spike}

\paragraph{Spike time gradients}

The gradients for $d_\S$ can be calculated as follows
\begin{equation}
     \pderivs{l_{s,p,o}}{\stime{s}} =  \pderivs{l_{s,p,o}}{\sscore{s}{p}{o}} \pderivs{\sscore{s}{p}{o}}{d_\S} \pderivs{d_\S}{\stime{s}} \,, 
\end{equation}
with
\begin{subequations}
\begin{align}
    \pderivs{l_{s,p,o}}{\sscore{s}{p}{o}} &= \eta_{s,p,o} \cdot \sigma \left( \sscore{S}{p}{o} \cdot \eta_{s,p,o} \right)\,, \\
    \pderivs{\sscore{s}{p}{o}}{d_\S} &= \sign{\dist{\S}{s}{o} - \vrel{p}} \,, \\
    \pderivs{d_\S}{\stime{s}} &=  \sign{\spikediff{s}{o}} \,.
\end{align}
\end{subequations}
All other gradients can be obtained similarly.

\paragraph{Weight gradients}

The spike times of nLIF neurons can be calculated analytically by setting the membrane potential equal to the spike threshold $\thresh$, i.e., $u_{s,i}(t^*) \overset{!}{=} \thresh$: 
\begin{equation}
    t^* = \taus \ln \bigg(  \underbrace{\frac{\nlifsum{j}\, e^{t_j \slash \taus}}{\nlifsumth{j}}}_{= T^*} \bigg) \label{si:spiketime} \,.
\end{equation}
In addition, for a neuron to spike, three additional conditions have to be met \citep{mostafa2017supervised}:
\begin{itemize}
    \item the neuron has not spiked yet,
    \item the input is strong enough to push the membrane potential above threshold, i.e.,
    \begin{equation}
        \nlifsum{j} > \thresh \label{si:spikecond1} \,,
    \end{equation}
    \item the spike occurs before the next causal pre-synaptic spike $t_c$, i.e., there is no other pre-synaptic spike influencing $t^*$,
    \begin{equation}
        t^* < t_c \label{si:spikecond2} \,.
    \end{equation}
\end{itemize}
From this, we can calculate the gradient
\begin{subequations}
\begin{align}
\pderivs{t^*}{W_{s,ik}} &= \frac{\taus}{T^*} \cdot \pderivs{T^*}{W_{s,ik}} \\
&= \frac{\taus \heaviside{t^* - t_k}}{T^*} \bigg[\frac{e^{t_k \slash \taus}}{\nlifsumth{j}} \nonumber \\
&- \frac{T^*}{\nlifsumth{j}}  \bigg] \\
&= \frac{\taus \heaviside{t^* - t_k}}{\nlifsumth{j}} \left[ \exp\left(\frac{t_k - t^*}{\taus}\right)  - 1 \right] \,,
\end{align}
\end{subequations}
where we used that $T^* = \exp\left(\frac{t^*}{\taus}\right)$. 

\subsection[Suppl. B]{Simulation details}\label{si:details}

Data, code and further examples using the Countries data set are available on \url{https://github.com/dodo47/SpikE}.

\paragraph{Regularization of weights}

To ensure that all neurons in the embedding populations spike, we use the same regularization term $L_\delta$ as \citet{mostafa2017supervised}
\begin{equation}
    L_\delta = 
\begin{cases}
    \sum_{s,i} \delta  \cdot \left(\thresh - w_{s,i} \right)     & \text{if } w_{s,i} \leq \thresh \,, \\
    0              & \text{otherwise}\,,
\end{cases}
\end{equation}
with $w_{s,i} = \sum_{j} W_{s,ij}$.

\paragraph{Software implementation}

Simulations were done using Python 3.7.7 and PyTorch 1.6.0.
For gradient updates, we use the Adagrad optimizer with $\epsilon = 10^{-10}$.
The loss is averaged per mini-batch.

\paragraph{Experiments}

We compute filtered metrics, i.e., where other triples that are also known to be true are removed from the ranking list.
For SpikE(-S), stimulating spikes are randomly initialized from a uniform distribution in the interval $[t_0, t_\mathrm{max}]$ and weights from a normal distribution $\mathcal{N}\left(0.2, 1.0\right)$.
Both for TransE(-S) and SpikE(-S), other embeddings are randomly initialized from a normal distribution $\mathcal{N}\left(0.0, 1.0\right)$.
For SpikE-S, we reduce the learning rate after 36 epochs to improve convergence.
$\thresh = 1$ in all cases.
For TransE we use the same loss function as for SpikE.
The remaining simulation parameters are given in \cref{tab:exp1}.
\begin{table}[h]\renewcommand{\arraystretch}{1.2}\caption{Simulation parameters.}\label{tab:exp1}\centerline{\begin{tabular}{c|ccc}
	\hline\hline
	                       & TransE(-S) & SpikE      & SpikE-S  \\
	\hline
	dimension              & 20         & 20         & 20 \\
	learning rate          & 0.1        & 1.0        & 1.0 $|$ 0.1 \\
	batch size             & 50         & 50         & 50 \\
	neg. samples           & 2$\times$2 & 2$\times$2 & 2$\times$2 \\
	L2 reg.                & 0.0001     & 0          & 0 \\
	stim. neurons          & -          & 40         & 40 \\
	$\taus$                & -          & 0.5        & 0.5 \\
	$[t_0,t_\mathrm{max}]$ & -          & $[-1,1]$         & $[-3,3]$ \\
	$\delta$               & -          & 0.01       & 0.01     \\
	\hline\hline
\end{tabular}}
\end{table} 

For the results shown in \cref{fig:dataevents}, we use a mini-batch size of 100, dimension of 12 and $\delta = 0.001$.
MRR, hits@k and general scores for these parameters are shown in \nameref{si:additional}.

\FloatBarrier

\subsection[Suppl. C]{Gating with parrot neurons}\label{si:gating} 

Gating either as subject or object of an embedding population can be realized using two populations of parrot neurons that immediately transmit their input and are gated instead.
This further allows the evaluation of relations that target the same subject and object population.
\FloatBarrier
\subsection[Suppl. D]{Synchronizing subject and object population}\label{si:synchronize}
\begin{figure}[h]
    \centering
    \includegraphics[width=0.8\columnwidth]{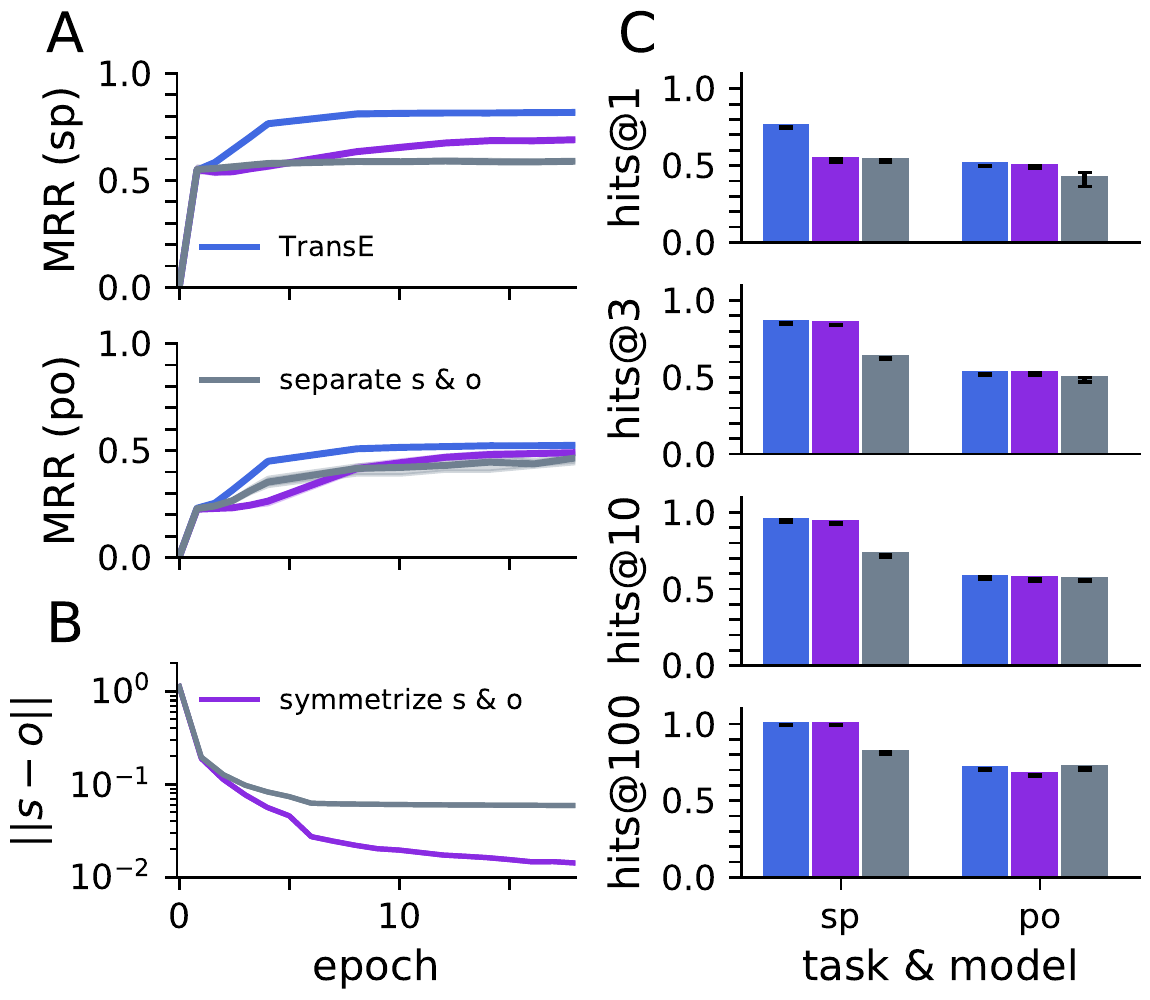}\vspace{-2mm}
	\caption{\textbf{(A)} MRR during training for TransE (blue), TransE with separate subject and object populations (gray) and TransE with separate populations and additional triple statements that enforce alignment (violet).
	\textbf{(B)} Strong alignment of subject and object populations can be observed when new triples are added during training.
	\textbf{(C)} Apart from the hits@1 score, adding alignment-preferring triples improves the performance.\vspace{-2mm}
	}	
	\label{fig:SIsymmetrize}
\end{figure}

If an entity is represented by distinct subject $s$ and object $o$ populations, these representations will differ after training despite representing the same entity.
By adding triples of the form \triple{s}{\textit{\#isIdenticalTo}}{o} and keeping $\vrel{\mathrm{isIdenticalTo}} = 0$, further alignment can be enforced (\cref{fig:SIsymmetrize}).
\FloatBarrier
\subsection[Suppl. E]{Different simulation parameters and static relations}\label{si:additional}

\begin{figure}[h]
    \centering
    \includegraphics[width=0.8\columnwidth]{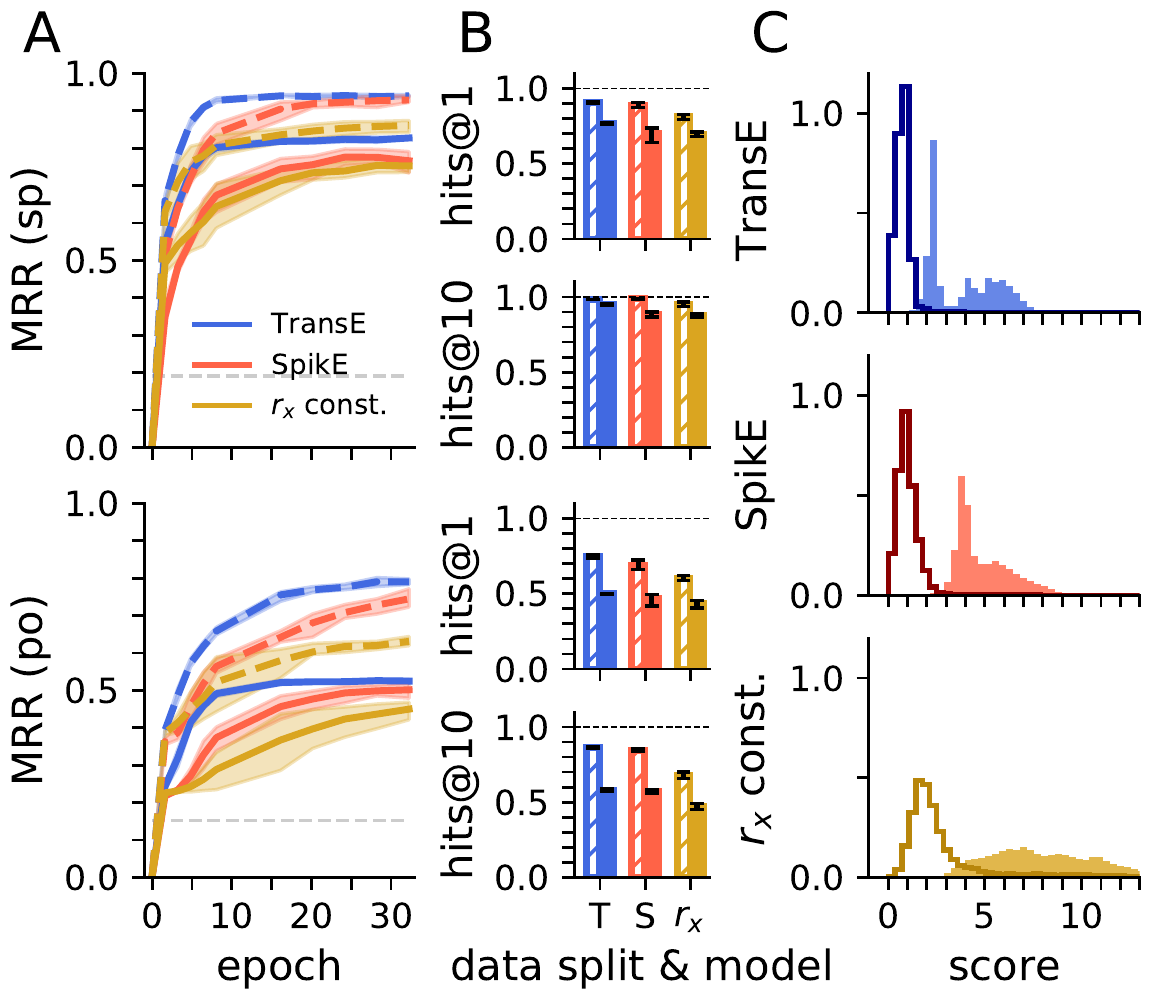}\vspace{-2mm}
	\caption{Same experiment as in \cref{fig:MRR}, but for a mini-batch size of 100, dimension of 12 and $\delta = 0.001$ (blue, red).
	Training still works if only node embeddings $\stime{x}$ are learned (yellow), i.e., relation embeddings $\vrel{x}$ are kept constant, but not vice versa, see gray dashed lines in A marking the best reached training MRR for that case.
	This highlights the importance of learning the spike times of the neural populations encoding entities in SpikE.\vspace{-4mm}
	}	
	\label{fig:SIlowerdim}
\end{figure}

\end{document}